\documentclass[10pt, journal, twocolumn]{IEEEtran}
\IEEEoverridecommandlockouts
\usepackage{cite}
\usepackage{booktabs}   % For professional-looking table lines (\midrule, \cmidrule)
\usepackage{tabularx}   % For fixed-width tables

\usepackage[pdftex]{graphicx}
\usepackage{float}
\usepackage{rotating}
\def\rot{\rotatebox}
\graphicspath{{../Figures/}}
\usepackage{todonotes}
\usepackage[cmex10]{amsmath}
\usepackage{soul}

\usepackage[hidelinks]{hyperref} % Already present and good for hiding link boxes
\usepackage{xr} 
\usepackage{subfiles}
\usepackage{algorithm}
\usepackage{algpseudocode}
\usepackage{amsmath}
\usepackage[acronym]{glossaries}
\newacronym{pca}{PCA}{Principal Component Analysis}
\newacronym{svd}{SVD}{Singular Value Decomposition}
\newacronym{nmf}{NMF}{Non-negative matrix factorization}
\newacronym{rpca}{RPCA}{Robust Principal Component Analysis}
\newacronym{orpca}{OR-PCA}{Online Robust Principal Component Analysis}
\newacronym{pcp}{PCP}{Principal Component Pursuit}
\newacronym{hp}{HP}{Hadamard Product}
\newacronym{ir}{IR}{implicit regularization}
\newacronym{ev}{EV}{Expressed Variance}
\newacronym{mgd}{MGD}{Momentum Gradient Descent}
\newacronym{ao}{AO}{alternating optimization}
\newacronym{omwrpca}{OMW-RPCA}{online moving window \acrshort{rpca}}
\newacronym{tforpca}{TF-ORPCA}{Tuning-Free \acrshort{orpca}}
\usepackage{array}
\usepackage{subcaption}
\usepackage{fixltx2e}
\usepackage{stfloats}
\usepackage{multirow}
\usepackage{array}
\usepackage{easyReview}
\usepackage{amsfonts}
\newcommand{\figref}[1]{Fig. \ref{#1}}
\AtBeginDocument{%
  \renewcommand\tablename{TABLO}
}

\AtBeginDocument{%
  \renewcommand\abstractname{Abstract}
}

	\usepackage[utf8]{inputenc} % Kullanılan encodinge göre utf8 yerine latin5 de yazılabilir.
	\usepackage[T1]{fontenc}

\begin{document}

% \IEEEpubid{\makebox[\columnwidth]{978-1-5386-1501-0/18/\$31.00 \copyright\ 2018 IEEE\hfill}
% \hspace{\columnsep}\makebox[\columnwidth]{}}

\title{Tuning-Free Online Robust Principal Component Analysis through Implicit Regularization}

%\author{{{Lakshmi Jayalal}\;\;\;{Gokularam Muthukrishnan}\;\;\;{Sheetal Kalyani}}}
% \author{
% 	\IEEEauthorblockN{%
% 		{Lakshmi Jayalal}, %
% 		$\,\ $%
% 		{Gokularam Muthukrishnan}, %
% 		$\,\ $%
% 		{Sheetal Kalyani}
%   }
% 	\IEEEauthorblockA{\textit{Department of Electrical Engineering,} \\
% 		\textit{Indian Institute of Technology Madras,}%\\
% 		\textit{
% 			Chennai - 600036, %INDIA. \\
% 			India.
% 		}
% 		\\
% 		e-mail: \{ee19d751@smail, ee17d400@smail, skalyani@ee\}.iitm.ac.in}
% }

\author{
	\IEEEauthorblockN{
		Lakshmi Jayalal\textsuperscript{*}, Gokularam Muthukrishnan\textsuperscript{*}, Sheetal Kalyani\textsuperscript{*}
		\thanks{\noindent\textsuperscript{*}The authors are with the Department of Electrical Engineering, Indian Institute of Technology Madras, Chennai 600036, India (%
        % e-mail: \texttt{ee17d400@smail.iitm.ac.in}; \texttt{ee23d001@smail.iitm.ac.in}; \texttt{skalyani@ee.iitm.ac.in}
        e-mail: \{ee19d751@smail, ee17d400@smail, skalyani@ee\}.iitm.ac.in%
        )
        }}
}



\maketitle

\begin{abstract}
 The performance of \acrfull{orpca} technique heavily depends on the optimum tuning of the explicit regularizers. This tuning is dataset-sensitive and often impractical to optimize in real-world scenarios. We aim to remove the dependency on these tuning parameters by using implicit regularization. 
To this end, we develop an approach that integrates implicit regularization properties of various gradient descent methods to estimate sparse outliers and low-dimensional representations in a streaming setting—a non-trivial extension of existing techniques.
A key novelty lies in the design of a new parameterization for matrix estimation in \acrshort{orpca}.
 Our method incorporates three different versions of modified gradient descent that separate but naturally encourage sparsity and low-rank structures in the data.
 Experimental results on synthetic and real-world video datasets demonstrate that the proposed method, namely, \acrfull{tforpca}, outperforms existing \acrshort{orpca} methods. \acrshort{tforpca} makes it more scalable for large datasets.

\boldmath
\end{abstract}



% \fi

\section{Introduction}
\acrfull{rpca} \cite{menon2019structured} is an outlier-resilient linear dimensionality reduction method. Several \acrshort{rpca} techniques 
have been proposed \cite{candes2011robust,fan2019factor,mateos2012robust}; with a detailed overview provided in \cite{vaswani2018robust}. In online settings where data arrive sequentially \cite{zhan2016online}, \acrfull{orpca} recursively obtains principal components \cite{feng2013online},\cite{li2022tensor}, reducing the memory footprint and improving efficiency by processing data as it is acquired. Variants like \acrfull{omwrpca} use a sliding window \cite{xiao2019online},\cite{zhang2024moving} for non-stationary environments.\cite{pulpito2023saliency} demonstrates how \acrshort{omwrpca} can be applied to real-world moving target detection. 
Other online approaches exist, such as GRASTA \cite{he2012incremental}, 
 robust subspace tracking \cite{narayanamurthy2018provable}, and residual-based \acrshort{orpca} \cite{zhu2022residual}. However, tuning regularization parameters poses significant practical hurdles for its widespread deployment. In this paper, we demonstrate that the dependency on regularization parameters can be alleviated through \acrfull{ir} methods \cite{zhang2021understanding,prechelt2002early,ziyin2024implicit,gunasekar2017implicit,vaskevicius2019implicit,wang2023implicit,li2023implicit}.

\acrshort{ir} leverages implicit biases in carefully designed optimization algorithms to enhance model generalization without %the need for 
explicit regularization terms. 
For instance, gradient descent often converges toward well-generalizing solutions despite multiple local minima in overparameterized problems, acting as a form of IR \cite{zhang2021understanding}, an effect further evident in techniques like early stopping and noise introduced by stochastic gradient descent that help avoid overfitting and improve generalization \cite{prechelt2002early,ziyin2024implicit}. 
Interestingly, even the choice of optimization algorithm introduces \acrshort{ir}. 
For example, in unconstrained, underdetermined least-squares problems, gradient descent reaches the minimum Euclidean norm solution \cite{gunasekar2017implicit}, while its parameterized form with early stopping can lead to minimal $\ell_1^{}$-norm solutions under certain conditions \cite{vaskevicius2019implicit}. 
We introduce \acrfull{tforpca}, a novel framework that integrates tailored \acrshort{ir} techniques for each subproblem (including a new algorithm for the subspace basis, adapted sparse IR, and early-stopped \acrfull{mgd}), 
efficiently managing streaming data with sparse outliers by removing dependency on the explicit regularization parameter. Experimental validation on synthetic and real-world datasets demonstrates that \acrshort{tforpca} outperforms existing \acrshort{orpca} techniques, achieving superior low-rank and outlier recovery by leveraging \acrshort{ir} and fixed hyperparameters, as opposed to traditional \acrshort{orpca} methods.

\textbf{Basic notations}: 
Bold, lowercase letters denote vectors; bold, uppercase for matrices. $\mathbf{X}^{\top}_{}$ is the transpose of $\mathbf{X}$. Column $\mathbf{z}_i^{}$ of the data matrix $\mathbf{Z}$ is the $i$-th data sample. $\|\mathbf{X}\|_F^{}$, $\|\mathbf{X}\|_\star^{}$, and, $\|\mathbf{x}\|_p^{}$ represent Frobenius, nuclear and  $\ell_p^{}$ norm, % 
respectively. $^\odot_{}$ and $\odot$ denote Hadamard power and Hadamard product, respectively.  
$\mathbf{1}_w^{}$ is a $w$-sized row vector of ones. 
\section[OR-PCA]{\acrshort{orpca}}
For the observed data $\mathbf{Z} = \mathbf{X}+\mathbf{E}$, where a sparse outlier $\mathbf{E}$ corrupts $\mathbf{X}$, the \acrshort{pcp} optimization problem is:
\begin{align}
	\min _{\mathbf{X}, \, \mathbf{E}} \ \frac{1}{2}\|\mathbf{Z}-\mathbf{X}-\mathbf{E}\|_F^2+\lambda_1\|\mathbf{X}\|_*+\lambda_2\|\mathbf{E}\|_1,
	\label{ORPCA_PCPEqn}
\end{align}
where $\mathbf{Z}, \mathbf{X}, \mathbf{E} \in \mathbb{R}^{p\times n}$ and $\lambda_1, \lambda_2$ are the explicit parameters. 

For sequential data, \acrshort{orpca} leverages bilinear decomposition of the low-rank matrix  as $\mathbf{X} = \mathbf{L} \mathbf{R}^\top$, where $\mathbf{L} \in \mathbb{R}^{p \times r}$ is the basis for the rank-$r$ subspace of $\mathbf{X}$ and rows of $\mathbf{R} \in \mathbb{R}^{n \times r}$ provide the corresponding coefficients. This provides a variational characterization of the nuclear norm as the infimum of the Frobenius norms of its factors when $r$ is known \cite{recht2010guaranteed}:
\begin{equation}
    \|\mathbf{X}\|_*=\inf _{\mathbf{L} \in \mathbb{R}^{p \times r},\, \mathbf{R} \in \mathbb{R}^{n \times r}}\, \left\{\frac{1}{2}\big(\|\mathbf{L}\|_F^2+\|\mathbf{R}\|_F^2\big): \mathbf{X}=\mathbf{L} \mathbf{R}^{\top}_{}\right\} .
    \label{ORPCA_nuclearNormExpression}
\end{equation} 
In online data, let $\mathbf{z}_t^{} \in \mathbb{R}^{p \times 1}{}$ be the sample revealed at time $t$. For a slowly changing basis $\mathbf{L}$, the revealed sample can be expressed as $\mathbf{z}_t^{} = \mathbf{L}\mathbf{r}_t^{\top} + \mathbf{e}_t^{}$, where $\mathbf{r}_t^{}\in \mathbb{R}^{1 \times r}_{}$ is its coefficient with respect to $\mathbf{L}$ and the corresponding sparse outlier $\mathbf{e}_t^{}\in \mathbb{R}^{p \times 1}_{}$. 
This leads to the online reformulation:
\begin{equation}
    \min _{\mathbf{L},\,\{\mathbf{r}_t^{}\},\, \{\mathbf{e}_t^{}\}} \frac{1}{n}\sum_{t=1}^n f(\mathbf{r}_t^{}, \mathbf{e}_t^{}, \mathbf{L}, \mathbf{z}_t^{})+ \frac{\lambda_1}{2n}\|\mathbf{L}\|_F^2, \label{eq:ORPCA_OnlineEqn}
\end{equation}
where
$f(\mathbf{r}_t^{}, \mathbf{e}_t^{}, \mathbf{L}, \mathbf{z}_t^{}):=\mathcal{L}_1\left(\mathbf{z}_t^{}, \mathbf{Lr}_t, \mathbf{e}_t^{}\right)+\frac{\lambda_1}{2}\|\mathbf{r}_t^{}\|_2^2 + \lambda_2 \|\mathbf{e}_t^{}\|_1
$ is the loss function and $\mathcal{L}_1\left(\mathbf{z}_t^{}, \mathbf{x}_t, \mathbf{e}_t^{}\right) := \frac{1}{2}\|\mathbf{z}_t^{}-\mathbf{x}_t-\mathbf{e}_t^{}\|_F^2$, the loss in data fidelity. \acrshort{orpca} 
employs alternating minimization to iteratively update $P_1^{}$, $P_2^{}$, and $P_3^{}$ at each step $t$.
 \begin{align}
  &P_1^{} := \min_{\mathbf{e}_t^{} \in \mathbf{R}^{p \times 1}} \mathcal{L}_1\left(\mathbf{z}_t^{},\mathbf{L}\mathbf{r}_t^{\top}, \mathbf{e}_t^{}\right)+\lambda_2\|\mathbf{e}_t^{}\|_1  \mid \left\{\mathbf{L}, \mathbf{r}_t^{}\right\} \label{eq:ORPCA_SolveE}, \\
  &P_2^{}:= \min_{\mathbf{r}_t^{} \in \mathbb{R}^{1 \times r}} \mathcal{L}_1\left(\mathbf{z}_t^{}, \mathbf{L}\mathbf{r}_t^{\top}, \mathbf{e}_t^{}\right)+\frac{\lambda_1}{2}\|\mathbf{r}_t^{}\|_2^2 \mid \left\{\mathbf{L}, \mathbf{e}_t^{}\right\} \label{eq:ORPCA_Eqnr},\\
  &P_3^{}:= \min_{\mathbf{L} \in \mathbf{R}^{p \times r}}\mathcal{L}_1\left(\mathbf{z}_t^{}, \mathbf{L}\mathbf{r}_t^{\top}, \mathbf{e}_t^{}\right) + \frac{\lambda_1}{2}\|\mathbf{L}\|_F^2 \mid \left\{\mathbf{e}_t^{} , \mathbf{r}_t^{}\right\} \label{eq:ORPCA_EqnL}.
\end{align}
The quality of the estimates depends on the proper setting of $\lambda_1$ and $\lambda_2$. These require data-dependent tuning via computationally expensive methods like grid search or cross-validation, often failing to generalize well to unseen data. 
\section[Tuning-free OR-PCA]{Tuning-free \acrshort{orpca}}
\begin{algorithm}
	\caption{Tuning-Free \acrshort{orpca}}
	\label{alg:ORPCA_AlgorithmPureIR}
	\hspace*{\algorithmicindent} \textbf{Input} $\mathbf{Z} = {\mathbf{z}_1^{}, \mathbf{z}_2^{}, \cdots, \mathbf{z}_T^{}}$, $n$ - number of samples, $r$ - intrinsic rank, $p$ - ambient dimension, $\mathbf{L}_0^{}, \mathbf{r}_0^{} , \mathbf{e}_0^{}$, $\mathbf{g}_0^{} \in \mathbb{R}^{p \times 1}, \mathbf{v}_0^{} \in \mathbb{R}^{p\times r}$ - initial solutions, $T_a^{} \forall a \in\{r,e,L\}$ - number of epochs for each parameter $\mathbf{r}_t^{}, \mathbf{e}_t^{}$ and $\mathbf{L}$, $T_0$ - max iterations for \acrlong{ao}.
	\begin{algorithmic}[1]
		\For {$t = 1$ to $n$}
		\State Reveal $\mathbf{z}_t^{}$ 
		\State $\mathbf{r}_0 = \mathbf{r}_{t-1}^{}, \mathbf{e}_0 = \mathbf{e}_{t-1}^{}, \mathbf{e}_t^{} = \mathbf{e}_{t-1}^{}$
		\Repeat
		\State $\mathbf{r}_t^{} \gets \texttt{MGD} (\mathbf{z}_t^{}-\mathbf{e}_t^{}, T_r)$
		\State $\mathbf{e}_t^{} \gets $ \texttt{HPGrad} ($\mathbf{z}_t^{}- \mathbf{L}_{t-1}^{}\mathbf{r}_t^{}, T_e$)
		\State $\epsilon \gets \text{max}\left\{\frac{\|\mathbf{r}_t^{} - \mathbf{r}_0\|_2^{}}{\|\mathbf{z}_t^{}\|{_2^{}}}, \frac{\|\mathbf{e}_t^{} - \mathbf{e}_0\|_2^{}}{\|\mathbf{z}_t^{}\|_2^{}}\right\}$
		\State $k \gets k + 1$
		\Until{$\epsilon < 10^{-3}$ or $k = T_0$}
		\State $\mathbf{L} =$  \texttt{HPGroupGrad}$\left(\mathbf{z}_t^{}- \mathbf{e}_t^{}, \mathbf{r}_t^{}, T_L, \mathbf{L}\right)$
		\State $\mathbf{R}[t, :] = \mathbf{r}_t^{}$; $\mathbf{E}[:, t] = \mathbf{e}_t^{\top}$
		\EndFor
		\State \textbf{return}: $\mathbf{L}$, $\mathbf{R}$, $\mathbf{E}$	
	\end{algorithmic}
\end{algorithm}
We utilize \acrshort{ir} methods to individually address problems $P_1^{}$, $P_2^{}$, and $P_3^{}$, thus circumvent tuning $\lambda_1$ and $\lambda_2$. Following \cite{feng2013online}, we alternately optimize $P_1^{}$ and $P_2^{}$ to estimate $\mathbf{e}_t^{}$ and $\mathbf{r}_t^{}$, based on the previous $\mathbf{L}$ estimate, until convergence. $\mathbf{e}_t^{}$ and $\mathbf{r}_t^{}$ then optimize $P_3^{}$, updating $\mathbf{L}$. This process relies on the most recent estimates for each optimization problem. Crucially, adapting \acrshort{ir} techniques for these specific subproblems within the \acrshort{orpca} framework requires a novel approach. The optimization for each sub-problem employs specific early stopping criteria, crucial for achieving \acrshort{ir}. The precise stopping rule details for each algorithm are provided in the Appendix.
The overall approach is summarized in Algorithm \ref{alg:ORPCA_AlgorithmPureIR}. In the following subsections, we discuss the solutions for each of these sub-optimization problems.
 \subsection[Estimating the sparse outlier e]{Estimating the sparse outlier $\mathbf{e}_t^{}$}
 \label{sec:ir_e}
To estimate the sparse outlier $\mathbf{e}_t^{}$ (optimization problem $P_1^{}$ (equation \eqref{eq:ORPCA_SolveE})), we adopt a modified gradient descent approach, as in \cite{vaskevicius2019implicit, li2021implicit}, which efficiently optimizes $\ell_1^{}$-norm regularized solution for \acrshort{orpca}.
This method involves decomposing the optimization parameter $\mathbf{e}_{}^{}$ in each iteration $i$. Then, these decomposed parameters are iteratively updated, resulting in a multiplicative update (lines 4, 5 of Algorithm \ref{alg:OptimizationOfe}). This decomposition inherently drives many non-significant components towards $0$, thereby implicitly inducing sparsity and eliminating the dependency on the parameter $\lambda_2^{}$. The approach is summarized in Algorithm \ref{alg:OptimizationOfe}, where the residual $\Delta$ updates the decomposed parameters $\mathbf{m}$ and $\mathbf{n}$. 
 \begin{algorithm}
	\caption{\texttt{HPGrad}$(\tilde{\mathbf{z}}_t, T_e)$}
	\label{alg:OptimizationOfe}
	\begin{algorithmic}[1]
		\State \textbf{Initialization}: $\alpha = 10^{-5}_{}$, $\eta = 5\times 10^{-3}_{}$, $\mathbf{m} = \alpha\mathbf{1}_p$, $\mathbf{n} = \alpha\mathbf{1}_p$, $\mathbf{e} = \mathbf{0}_p^{}$.
		\For{$i = 0$ to $T_e^{}-1$}
        \State $\Delta \gets \frac{4}{p}(\tilde{\mathbf{z}}_t^{} - \mathbf{e})$
        \State $\mathbf{m} \gets \mathbf{m} \odot (1 - \eta \Delta); \quad \mathbf{n} \gets \mathbf{n} \odot (1 + \eta \Delta)$
		\State $\mathbf{e} \gets \mathbf{m}_{}^{\odot2} - \mathbf{n}_{}^{\odot2}$
		\EndFor
%        \State $\mathbf{e}_t^{} \gets \mathbf{e}$
        \State \Return $\mathbf{e}$
	\end{algorithmic}
\end{algorithm}
\subsection[Estimating the Coefficient Vector r]{Estimating the Coefficient Vector $\mathbf{r}_t^{}$}
$P_2^{}$ given in equation \eqref{eq:ORPCA_Eqnr} is a ridge regression problem 
where $\lambda_1$ controls the $\ell_2$-norm   to estimate $\mathbf{r}_t^{}$. 
We use vanilla \acrshort{mgd} to estimate $\mathbf{r}_t$, motivated by its theoretical connection to $\ell_2$-regularization, established and explored in \cite{wang2023implicit}.
This suggests that \acrshort{mgd} is a viable tuning-free alternative to $\ell_2^{}$-regularization loss.  The early stopping time $T_r$ of vanilla \acrshort{mgd} acts as an \acrshort{ir} parameter, related to the ridge regression tuning parameter by $\lambda = \frac{2}{t^2}$. By setting $T_r$, we implicitly control the $\ell_2$-regularization, obviating $\lambda_1$ from \eqref{eq:ORPCA_Eqnr}. The momentum parameter $\mu$ is set consistently to $0.9$. The approach is outlined in Algorithm \ref{alg:ORPCA_EstimateR}, where the residual is computed, used to update the velocity of $\mathbf{r}$, then subsequently $\mathbf{r}$ itself. 

\begin{algorithm}
	\caption{\texttt{MGD}$(\tilde{\mathbf{z}}_t^{}, \mathbf{L}, T_r^{})$}\label{alg:ORPCA_EstimateR}
	\begin{algorithmic}[1]
		\State \textbf{Initialization}: $\alpha = 10^{-5}_{}$, $\eta = 5\times 10^{-3}_{}$, $\mu = 0.9$, $\mathbf{r} = \alpha\mathbf{1}_r$, $\mathbf{v}_r = \mathbf{0}_r^{}$
		\For{$i = 0$ to $T_r-1$}
		\State $\Delta \gets \frac{4}{p}(\tilde{\mathbf{z}}_t - \mathbf{L}\mathbf{r})$
		\State $\mathbf{v}_r \gets \mu \mathbf{v}_r - \eta \Delta$;\quad $\mathbf{r} \gets \mathbf{r} + \mathbf{v}_r$
		\EndFor
		\State \Return  $\mathbf{r}$
	\end{algorithmic}
\end{algorithm}
This implicitly eliminates $\lambda_1$ from \eqref{eq:ORPCA_Eqnr}. 
To eliminate the dependency on the parameter $\lambda_1$ from equation \eqref{eq:ORPCA_OnlineEqn}, we estimate the basis 
implicitly in the next subsection.

\subsection[Estimating the subspace basis L]{Estimate the subspace basis $\mathbf{L}$}

Estimating $\mathbf{L}$ requires minimization of the loss function penalized by the Frobenius norm of $\mathbf{L}$ (optimization problem $P_3^{}$ (equation \eqref{eq:ORPCA_EqnL})). A standard matrix factorized gradient descent method \cite{gunasekar2017implicit} for matrix completion is insufficient, as it tends to converge to the nuclear norm of the matrix  and is thus not directly applicable here. Addressing this challenge is a key novel aspect of \acrshort{tforpca}. 
The proposed Algorithm \ref{alg:ORPCA_EstimateL} employs a unique and novel reparameterization $\mathbf{L} = \mathbf{g}^{\odot2}_{}\mathbf{1}_r^{}\odot\mathbf{V}$, where $\mathbf{g}\in \mathbb{R}^{p \times 1}{}$, and $\mathbf{V} \in \mathbb{R}^{p \times r}{}$. The intuition behind parameterization is that it separates the control of the magnitude in the row $\mathbf{g}$ and the direction of the basis in the element $\mathbf{V}$. 
This separation allows for refined control over the low-rank structure by implicitly balancing the magnitude of rows and the directional components of the basis, thereby mimicking the effect of the Frobenius norm regularization.
The associated updates (lines 4-6) implicitly manage the basis structure, aiming to mimic the effect of $\|\mathbf{L}\|_F^2$ without the dependence on the tuning parameter $\lambda_1$ in equation \eqref{eq:ORPCA_EqnL}.
As with the previous algorithms, each element is individually updated at each iteration; the residual is used to update the parameters $\mathbf{g}$ and $\mathbf{V}$, which then calculate $\mathbf{L}$. 
While rigorous theoretical guarantees for this novel algorithm are part of an ongoing investigation and future work, we emphasize its empirical effectiveness in the following section that incorporates this novel reparameterization.

\begin{algorithm}
	\caption{\texttt{HPGroupGrad}$(\mathbf{z}_t^{}, \mathbf{r}_t^{}, T_L^{}, \mathbf{L})$}
	\label{alg:ORPCA_EstimateL}
	\begin{algorithmic}[1]
		\State \textbf{Initialization}: $\eta = 5\times 10^{-3}_{}$, $\mathbf{L}_0 = \mathbf{L}$, $i = 0$
        \Repeat
		\State $\Delta \gets \left(\mathbf{z}_{t} - \mathbf{L}\mathbf{r}^{\top}_{t}\right)\mathbf{r}_{t}$
		\State $\mathbf{g} \gets \mathbf{g} - \frac{\eta}{p}\left(\Delta\odot(\mathbf{g}\mathbf{I}_r)\odot\mathbf{V}\right)\mathbf{I}_r^{\top}$
		\State $\mathbf{V} \gets \mathbf{V} - \frac{\eta}{p}\Delta \odot(\mathbf{g}_{}\mathbf{I}_r)^{\odot2}$
		\State $\mathbf{L} \gets (\mathbf{g}_{i}^{\odot2}\mathbf{I}_r)\odot\mathbf{V}$
		\State $i \gets i+1 $
        \Until {
        $i = T_L$}
		\State $\mathbf{g}_0 \gets \mathbf{g}; \mathbf{v}_0 \gets \mathbf{v}$
		\State \Return $\mathbf{L}$
	\end{algorithmic}
\end{algorithm}

\begin{figure*}[ht]
    \centering
    % Use a tabular environment for a perfectly aligned grid
    % l: left-aligned column for the dataset name
    % c: centered columns for the images
    
    \setlength{\tabcolsep}{0.5pt}
    \begin{tabularx}{\linewidth}{@{}l@{\hspace{0.5mm}} *{7}{c}@{}}
        \toprule
        % --- Main Column Headers ---
        \addlinespace[1mm]
        & \textbf{Original} & \multicolumn{2}{c}{\textbf{ Method}} & \multicolumn{2}{c}{\textbf{\acrshort{orpca}}} & \multicolumn{2}{c}{\textbf{\acrshort{omwrpca}}} \\
        \cmidrule(lr){3-4} \cmidrule(lr){5-6} \cmidrule(lr){7-8}
        % --- Sub-Headers for L and E components ---
        & & \textbf{($\mathbf{L}$)} & \textbf{($\mathbf{E}$)} & \textbf{($\mathbf{L}$)} & \textbf{($\mathbf{E}$)} & \textbf{($\mathbf{L}$)} & \textbf{($\mathbf{E}$)} \\
        \midrule

        % --- PETS2006 Row ---
        \rot{90}{\textbf{PETS2006}} &
        \includegraphics[width=0.1380\linewidth, height = 2cm]{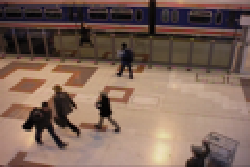} &
        \includegraphics[width=0.1380\linewidth, height = 2cm]{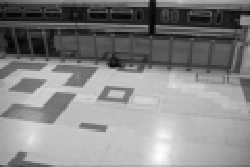} &
        \includegraphics[width=0.1380\linewidth, height = 2cm]{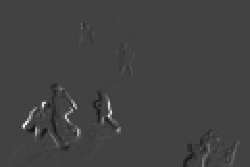} &
        \includegraphics[width=0.1380\linewidth, height = 2cm]{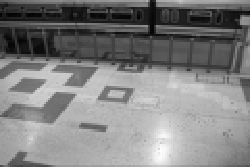} &
        \includegraphics[width=0.1380\linewidth, height = 2cm]{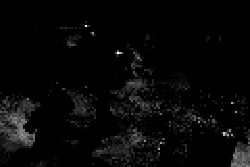} &
        \includegraphics[width=0.1380\linewidth, height = 2cm]{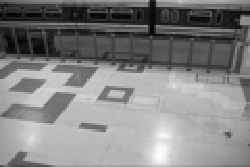} &
        \includegraphics[width=0.1380\linewidth, height = 2cm]{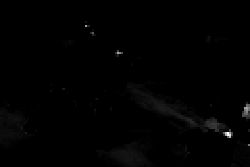} \\

        % --- Pedestrians Row ---
        \rot{90}{\textbf{Pedestrians}} &
        \includegraphics[width=0.1380\linewidth, height = 2cm]{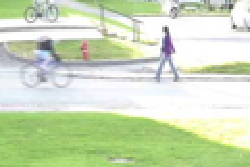} &
        \includegraphics[width=0.1380\linewidth, height = 2cm]{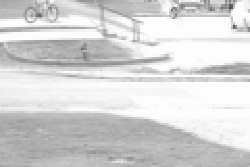} &
        \includegraphics[width=0.1380\linewidth, height = 2cm]{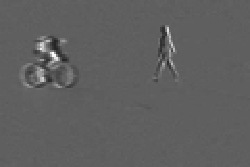} &
        \includegraphics[width=0.1380\linewidth, height = 2cm]{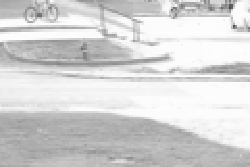} &
        \includegraphics[width=0.1380\linewidth, height = 2cm]{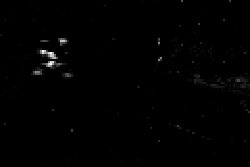} &
        \includegraphics[width=0.1380\linewidth, height = 2cm]{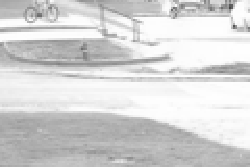} &
        \includegraphics[width=0.1380\linewidth, height = 2cm]{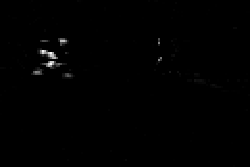} \\

        % --- Bungalows Row ---
        \rot{90}{\textbf{Bungalows}} &
        \includegraphics[width=0.1380\linewidth, height = 2cm]{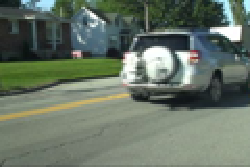} &
        \includegraphics[width=0.1380\linewidth, height = 2cm]{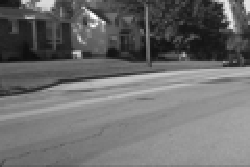} &
        \includegraphics[width=0.1380\linewidth, height = 2cm]{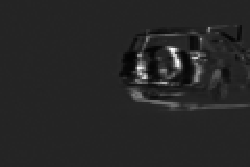} &
        \includegraphics[width=0.1380\linewidth, height = 2cm]{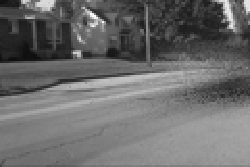} &
        \includegraphics[width=0.1380\linewidth, height = 2cm]{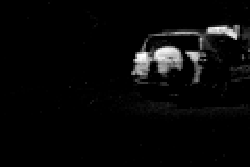} &
        \includegraphics[width=0.1380\linewidth, height = 2cm]{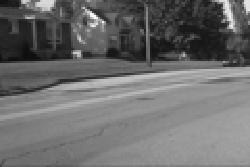} &
        \includegraphics[width=0.1380\linewidth, height = 2cm]{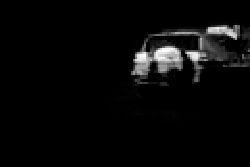} \\
        \bottomrule
    \end{tabularx}
    \caption{
    Video surveillance in various datasets (top to bottom: PETS2006, Pedestrians, and Bungalows).
    Columns (left to right) illustrate: col. $1$ $-$ Original image, col. $2\,\&\,3$ $-$ Low-rank background $(\mathbf{L})$ and sparse foreground $(\mathbf{E})$ components recovered by \acrshort{tforpca}, col. $4\, \& \, 5$ $-$ $\mathbf{L}$ and $\mathbf{E}$ recovered by \acrshort{orpca}, col. $6\, \& \, 7$ $-$ $\mathbf{L}$ and $\mathbf{E}$ recovered by \acrshort{omwrpca}}
    \label{fig:ORPCA_RealDataPedestrians479}
\end{figure*}

\section{Simulation Results}
\subsection{Synthetic Data Case Studies}

\begin{figure}[t]
    \centering
    \includegraphics[width=0.895\linewidth]{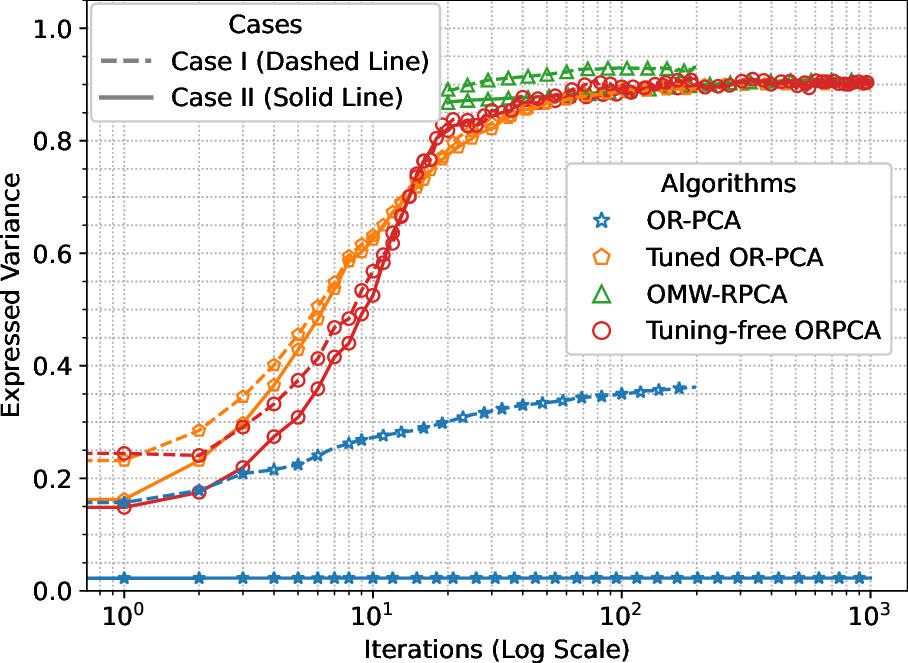}%
    \caption{Comparison of EV for four different \acrshort{orpca} methods.}
    \label{fig:ORPCA_SimulatedComparision}
\end{figure}

For synthetic evaluations, $\mathbf{X} = \mathbf{U}\mathbf{V}^{\top}_{}$ is generated with $n = 200$, ambient dimension $p = 80$, where $\mathbf{U}, \mathbf{V} \sim \mathcal{N}(0, \frac{1}{n})$, and the intrinsic dimension of the subspace spanned by $\mathbf{U}$ is $r = 10$. The observed samples are $\mathbf{Z} = \mathbf{X} + \mathbf{E}$, with $\mathbf{E} \sim \mathcal{U}(-1000, 1000)$, where a fraction $(\rho = 0.01)$ of its entries is non-zero. This configuration serves as \textbf{Case I} of our studies. For \textbf{Case II}, we use $n = 1000$, $p = 400$, $r = 10$, and $\rho = 0.01$. 
To compare the performance, we use \acrfull{ev} \cite{xu2010principal}. Higher \acrshort{ev} indicates better recovery of the underlying low-rank structure. For both \acrshort{orpca} and \acrshort{omwrpca} $\lambda_1, \lambda_2$ are set to $1/\sqrt{p}$, as designed by the corresponding authors. 
We also included results from carefully tuned parameters $(\lambda_1 = \lambda_2 = 1)$ to achieve optimal \acrshort{ev} after a grid search of values from $0$ to $1.2$ in increments of $0.1$.
For \acrshort{omwrpca}, the initial data points are reserved for rank estimation, which explains the absence of metrics in the initial iterations in \figref{fig:ORPCA_SimulatedComparision}. Figures plot the average \acrshort{ev} over 10 experiments.

The hyperparameters for the \acrshort{tforpca} are set with initial values are $m_0 = n_0 = r_0 = V_0 = 10^{-5}$, and $g_0 = 1$.
The learning rate is set to $\eta_r^{} =\eta_e^{} = \eta_g^{} = \eta_v^{}  = 5 \times 10^{-3}$. The number of epochs, $T_e^{}, T_r^{}$, and $T_L^{}$ are determined by Equations (7)-(9) in the Appendix. 
As \figref{fig:ORPCA_SimulatedComparision} demonstrates, all algorithms' performance steadily improves with more samples. Notably, the \acrshort{ev} of the \acrshort{tforpca} is consistently better than that of \acrshort{orpca}. We also observe that the \acrshort{tforpca} converges to a performance level comparable to that of carefully tuned \acrshort{orpca} and \acrshort{omwrpca} with window size $15$, without requiring any dataset-dependent parameter tuning. 
The plot for \acrshort{orpca} shows significant degradation in performance (\acrshort{ev}$\leq  0.4$), demonstrating that it is susceptible to regularization parameters. The high performance of tuned-\acrshort{orpca} and \acrshort{omwrpca} is attributed to the careful tuning of the data-dependent parameters $\lambda_1^{}, \lambda_2^{}$, and the window size. The better performance observed in \acrshort{tforpca} stems from the adaptive nature of the \acrshort{ir} introduced due to the different problem-specific reparameterizations used. 

\subsection{Ablation Studies: Hyperparameter Sensitivity}
\begin{figure}[t]
    \centering
    \includegraphics[width=0.895\linewidth]{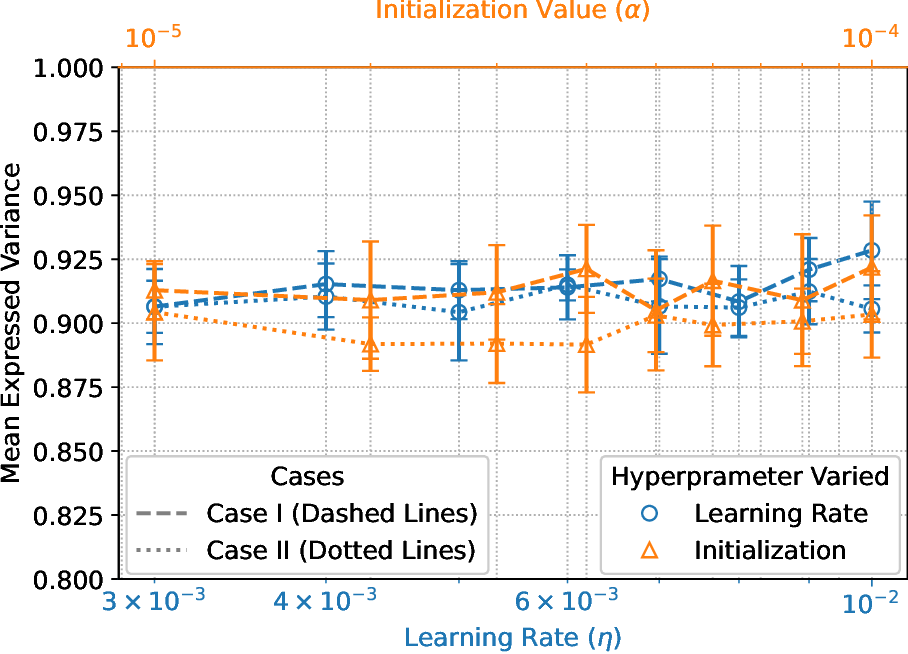}
    \caption{Ablation Study of Tuning-Free \acrshort{orpca}}
    \label{fig:ORPCA_AblationStudy}
\end{figure}

To evaluate the robustness of \acrshort{tforpca}, an ablation study was conducted on its hyperparameters. We specifically investigated the impact of the learning rate $(\eta)$ and the initialization value $(\alpha)$. We varied $\eta$ across a range of values for the learning rate analysis while the initialization values were set to a fixed $\alpha$. In contrast, for the initialization value analysis, $\alpha$ was varied with a fixed $\eta$. \figref{fig:ORPCA_AblationStudy} demonstrates the remarkable stability and insensitivity of the algorithm to variations in both the learning rate and initialization values within the evaluated ranges, suggesting inherent robustness. More detailed information on these ablation studies can be found in Appendix.


\subsection{Real Dataset}
Surveillance video is an excellent candidate for studying \acrshort{orpca}, where the background is the slow-changing low-rank component and the foreground corresponds to sparse outliers. 
A tuning-free approach is especially useful in these diverse, often unpredictable environments, as it eliminates the need for manual parameter adjustments. 
For comparison, the \acrshort{orpca} and \acrshort{omwrpca} baselines were run with their typical default parameters, $\lambda_1 = \lambda_2 = 1/\sqrt{p}$.
\figref{fig:ORPCA_RealDataPedestrians479} illustrates the low-rank matrix and the sparse outlier recovery for the PETS2006, Pedestrians, and Bungalows datasets from the change detection dataset. The hyperparameters are kept the same across all datasets: $\alpha = 10^{-5}_{}, \eta = 5\times 10^{-3}_{}$, and epochs ($T_e^{}, T_r^{}$, and $T_L^{}$) determined by Equations (7)-(9) in the Appendix. The images are resized from $240 \times 360$ to $48 \times 72$ to accommodate the algorithms.

The estimation of the low-rank matrix and the sparse outliers using \acrshort{tforpca} is clear and exhibits minimal shadowing in the recovered low-rank matrix for all datasets. The results for \acrshort{orpca} (columns 4 and 5 of \figref{fig:ORPCA_RealDataPedestrians479}) highlight limitations in its consistent performance across datasets. Specifically, while low-rank recovery (column 4) appears satisfactory for the Pedestrians dataset, considerable background shadowing is evident in the Bungalows dataset, particularly around the estimated vehicle.
The PETS2006 dataset exhibits undesirable gray patches in its low-rank recovery. Despite successful outlier extraction (column 5) in the Bungalows dataset, the quality of outlier recovery in the Pedestrians and PETS2006 datasets is compromised, with foreground objects appearing as significant noise rather than clear detections.

The results for \acrshort{omwrpca} (columns 6 and 7) demonstrate effective low-rank recovery and outlier detection for the Bungalows dataset and show proper low-rank estimation for the Pedestrians dataset. While the low-rank estimation for the PETS2006 dataset is generally good, it exhibits a minor estimation error spanning a few pixels. Furthermore, outlier detection for both the Pedestrians and PETS2006 datasets is poor, with a complete lack of distinct detections. Together, these observations for \acrshort{orpca} and \acrshort{omwrpca} strongly suggest that their performance is highly data sensitive, necessitating meticulous parameter tuning for each specific dataset.
\acrshort{tforpca} eliminates this need by using consistent hyperparameters across datasets, a robustness that is particularly advantageous in real-world scenarios where ground-truth or validation data for parameter selection are unavailable.
\section{Conclusion}

We introduced tuning-free \acrshort{orpca}, which leverages implicit regularization to eliminate the regularization parameters. Our approach employs three problem-specific \acrshort{ir} techniques to implicitly promote sparsity and a low-rank structure : (a) a modified gradient descent for an implicit $\ell_1$-regularizer, (b) an early-stopped momentum gradient descent for implicit $\ell_2$-regularizer, and (c) a novel reparameterization estimates the subspace basis that implicitly regularizes the Frobenius norm. Unlike traditional \acrshort{orpca}, which requires fine-tuning regularization parameters, our algorithm is relatively insensitive to the choice of its parameters, and hence, does not require extensive tuning. This enables \acrshort{orpca} to be applied to real-world applications, such as video surveillance, where traditional tuning-dependent approaches are often ineffective. Our method clearly extracts foreground and background components without the artifacts, such as shadowing or gray patches, often seen with traditional approaches.

\appendix

\section{Algorithm Details}
\label{sec:algorithm_details}
This section provides the early stopping criteria used in the algorithms presented in the main paper. 
    The alternating optimization loop in Algorithm 1 runs for a maximum of $T_0 = 50$ iterations. Each sub-optimization for $\mathbf{e}_t, \mathbf{r}_t$,  and $\mathbf{L}$ (Algorithms 2, 3, and 4) is performed for a fixed number of epochs, denoted as $T_e^{}$, $T_r^{}$, and $T_L^{}$, respectively. For algorithm 2, the maximum number of iterations, $T_e^{}$, is set according to the derivation in \cite{li2021implicit} as: 
    \begin{align}
        T_e^{} &=\frac{15}{32} n\log_2^{}\left(\frac{\max(|\tilde{\mathbf{z}}_t|) - \alpha^2_{}}{ \eta^2_{} \epsilon}\right)\label{eqn:ORPCA_Te},
    \end{align}  where $n$ is the number of samples, $\alpha$ is the initialization constant, $\eta$ is the learning rate, and $\epsilon$ is the expected accuracy between updates and the oracle value.  Similarly, for algorithms 3 and 4, the maximum number of iterations for $T_r^{}$ is set to: \begin{align}
        T_r^{} &= \frac{15}{32} r \log_2^{}\left(\frac{\max(|\tilde{\mathbf{z}_t^{}}|) - \alpha^2_{}}{ \eta^2_{} \epsilon}\right),\label{eqn:ORPCA_Tr}\\
        T_L^{} &= \frac{15}{32} nr \log_2^{}\left(\frac{\max(|\mathbf{z}_t^{}|) - \alpha^2_{}}{ \eta^2_{} \epsilon}\right). \label{eqn:ORPCA_Tl} 
    \end{align}  This duration is determined similarly to $T_e^{}$, considering the dimensions $n$ and $r$ of the matrices involved.

\section{Additional Experimental Results}
\label{sec:additional_results}

This section presents further experimental results that complement those reported in the main paper.

\subsection{Ablation Study: Sensitivity to hyperparameters}
\label{subsec:TFORPCA_ablation_init}

This section provides further details on the algorithm's hyperparameter sensitivity. The same synthetic datasets as in Cases I and II are utilized for the learning rate analysis. All initialization values were consistently set to $\alpha = 10^{-5}_{}$. A single learning rate was uniformly applied across all constituent subalgorithms 
% (Algorithms 2,3 and 4) 
during optimization. We varied $\eta$ across eight distinct values ranging from $3\times 10^{-3}_{}$ to $10^{-2}_{}$. Fig. 3 presents the mean \acrshort{ev} obtained from $10$ independent experiments for each tested learning rate and initialization value.

Fig. 3 distinctly shows that the algorithm's performance, in terms of the mean \acrshort{ev}, remains remarkably stable and largely insensitive to variations in both the learning rate and the initialization value within the evaluated range.  Specifically, for the initialization value, this insensitivity holds, provided the value is not excessively small, as extremely low initializations can lead to stagnation in gradient descent methods utilizing Hadamard reparameterization (e.g., Algorithm 2 for sparse noise estimation). For both studies, the y-axes were zoomed in for clarity, and the error bars depict the standard deviation of 10 experiments, further substantiating consistent performance. Collectively, these studies reveal a strong insensitivity of the algorithm to the precise choice of initialization values and learning rates within the evaluated ranges, suggesting inherent robustness.

\bibliography{main.bib}
\bibliographystyle{IEEEtran}
\end{document}